\documentclass[letterpaper, 10 pt, conference]{IEEEconf}  

\IEEEoverridecommandlockouts                              

\usepackage{graphicx} 
\usepackage{svg}
\usepackage{amsmath} 
\usepackage{amssymb}  

\title{\LARGE \bf
Tactile Active Inference Reinforcement Learning for Efficient Robotic Manipulation Skill Acquisition
}

\author{Zihao Liu$^{\dagger}$, Xing Liu$^{\dagger}$, Yizhai Zhang, Zhengxiong Liu and Panfeng Huang$^{*}$ 
\thanks{This work was supported in part by the National Key R\&D Program of China under Grant 2022ZD0117900, and the National Natural Science Foundation of China under Grant 62103334 and 62273280.}
\thanks{Zihao Liu, Xing Liu, Yizhai Zhang, Zhengxiong Liu, and Panfeng Huang(Corresponding author) are with the Research Center for Intelligent Robotics, School of Astronautics, Northwestern Polytechnical University, and National Key Laboratory of Aerospace Flight Dynamics, Northwestern Polytechnical University, Xi'an, China, 710072 e-mail: pfhuang@nwpu.edu.cn, xingliu@nwpu.edu.cn.}%
}

\begin{document}

\maketitle
\thispagestyle{empty}
\pagestyle{empty}

\begin{abstract}

Robotic manipulation holds the potential to replace humans in the execution of tedious or dangerous tasks. However, control-based approaches are not suitable due to the difficulty of formally describing open-world manipulation in reality, and the inefficiency of existing learning methods. Thus, applying manipulation in a wide range of scenarios presents significant challenges. In this study, we propose a novel method for skill learning in robotic manipulation called Tactile Active Inference Reinforcement Learning (Tactile-AIRL), aimed at achieving efficient training. To enhance the performance of reinforcement learning (RL), we introduce active inference, which integrates model-based techniques and intrinsic curiosity into the RL process. This integration improves the algorithm's training efficiency and adaptability to sparse rewards. Additionally, we utilize a vision-based tactile sensor to provide detailed perception for manipulation tasks. Finally, we employ a model-based approach to imagine and plan appropriate actions through free energy minimization. Simulation results demonstrate that our method achieves significantly high training efficiency in non-prehensile objects pushing tasks. It enables agents to excel in both dense and sparse reward tasks with just a few interaction episodes, surpassing the SAC baseline. Furthermore, we conduct physical experiments on a gripper screwing task using our method, which showcases the algorithm's rapid learning capability and its potential for practical applications.

\end{abstract}

\section{INTRODUCTION}

The manipulation of robots has broad application prospects, and it can replace humans in performing diverse tasks. Currently, robotic manipulation mostly relies on programmed instructions for control. Developers use their knowledge to design procedures and form task plans, and robots perform skilled behaviors based on position and force information \cite{tactile-servo}.
However, this approach has two drawbacks that limit its widespread use in reality. First, real-world scenarios are often unstructured, making it difficult to be described completely with standardized geometric shapes and physical properties. Second, human knowledge cannot foresee all situations that robots will encounter, making complex task planning that still has blind spots. These challenges have led to the current application of robotic manipulation being mostly limited to well-defined environments with fixed scenarios. 
To mitigate these two challenges, practitioners leverage tactile sensing to achieve a more comprehensive perception of the scene, and use reinforcement learning to obtain policies or plans for each state.
As manipulation tasks often involve contact forces, tactile information provides a more detailed description of the scene compared to pose information. Moreover, RL methods can adapt to a wide range of state configurations and learn spontaneously through exploration. However, traditional tactile sensing suffers from information sparsity \cite{haptic-touch}, and the RL paradigm also suffers from low data efficiency.

\begin{figure}[thpb]
      \centering
      \includegraphics[scale=0.25]{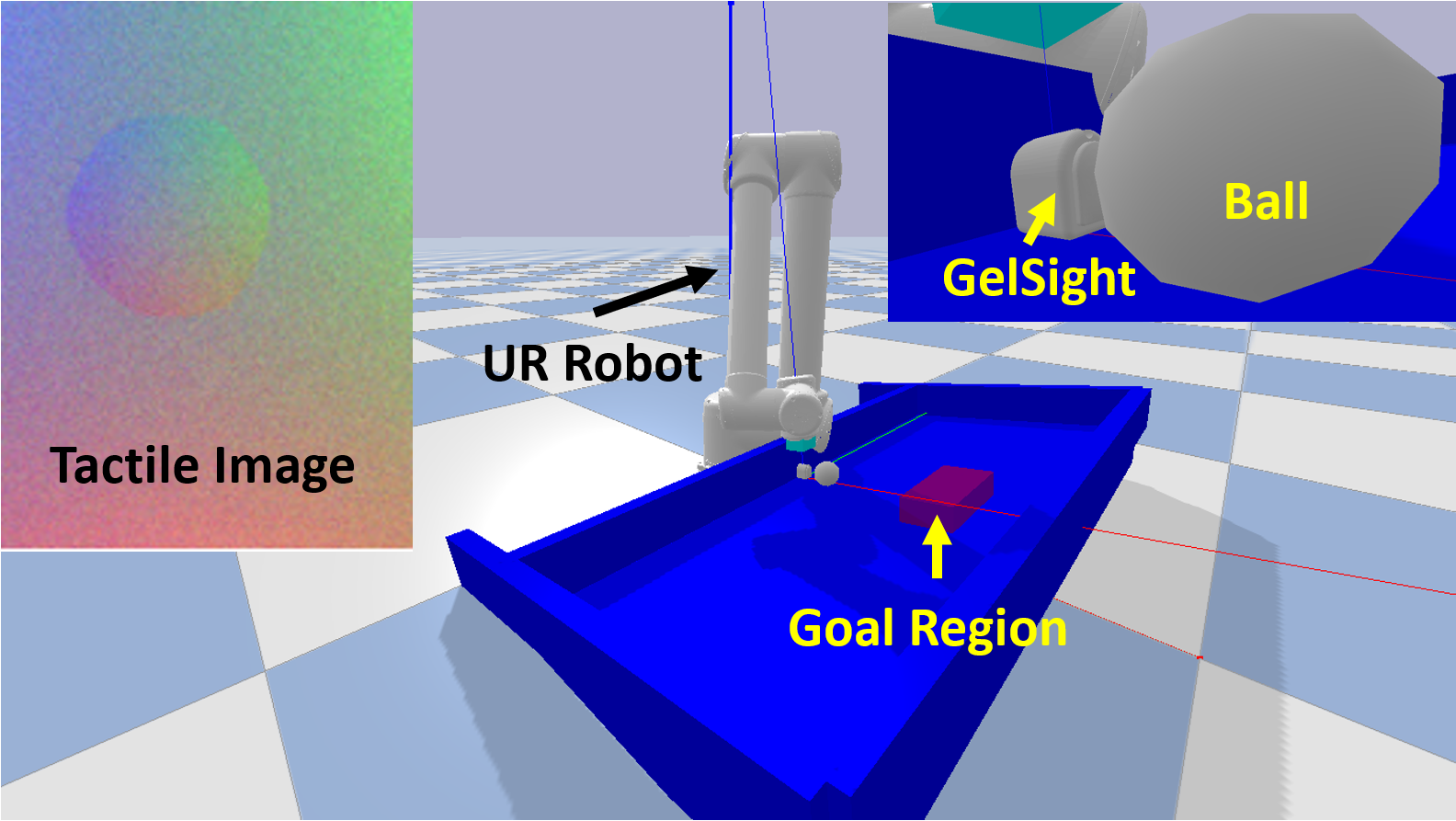}
      \caption{Object pushing on a slope: The UR5 manipulator is tasked with pushing a ball or box from the lower end of the slope to the upper red goal region. }
      \label{sim_scene}
    \end{figure}

Different from the commonly-used force/torque sensors or strain-based tactile sensors, vision-based tactile sensors such as GelSight \cite{GelSight} offer a new approach to obtaining tactile information by deforming a gel that contacts an object. 
In this work, we explore the integration of vision-based tactile sensors in robot manipulation skill learning to enhance learning efficiency. 
On the other hand, to address the data efficiency issue of vanilla RL, model-based RL \cite{muzero}\cite{dreamer} has been widely employed. Typically, model-based methods learn a state transition model of the agent's environment and use the model for offline or online planning and action execution. By decoupling the model learning from the policy learning, model-based methods indeed improve the utilization efficiency of sampled data. 
However, the issue of accumulated errors still exists, leading to poor performance of model-based RL in inappropriate reward context such as sparse reward, indicating the limitations of model-based RL in exploration. 
Therefore, to efficiently train robotic RL models, it is essential to combine data exploration with exploitation. And this approach can enhance the adaptability of models to sparse rewards, thereby reducing the difficulty of designing reward functions. Finally, this could increase the feasibility of learning robotic manipulation skills in real-world scenarios.

\begin{figure}[thpb]
      \centering
      \includegraphics[scale=0.04]{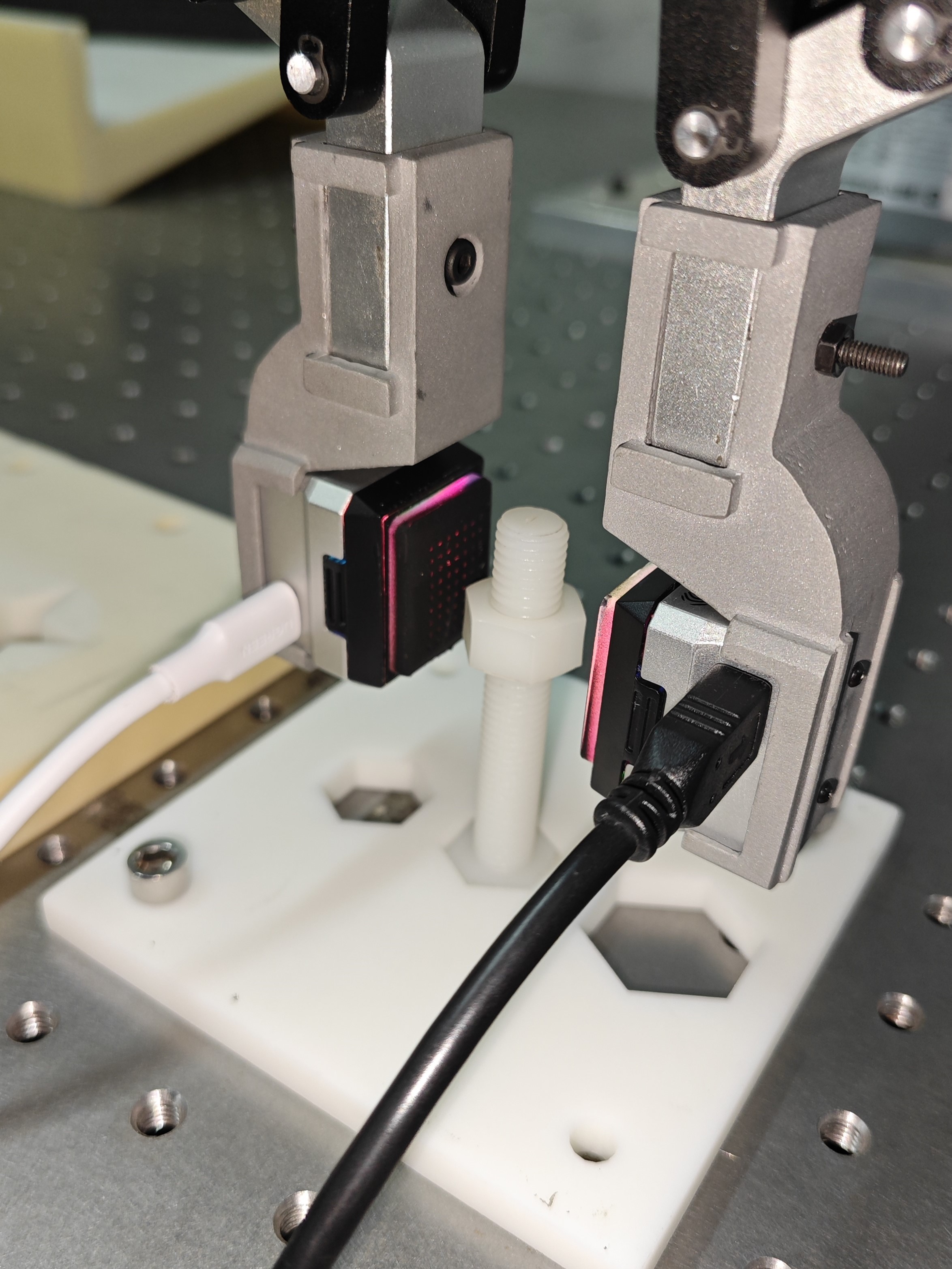}
      \caption{Robot screwing: The KUKA manipulator equipped with the gripper will adaptively hold the nylon nut and utilize tactile sensor feedback to tighten it.}
      \label{real_scene}
    \end{figure}

To summarize our contributions in this work:
\begin{itemize}
\item We incorporate vision-based tactile information into the state space of reinforcement learning, enhancing the perception of the manipulation task.
\item We introduce active inference reinforcement learning for acquiring robot manipulation skills, aiming to minimize free energy for increasing environmental awareness and achieving desired observations.
\item We validate our method for learning robotic manipulation skills through simulation and real-world experiments. In the simulation, we employ a manipulator to learn object pushing across a slope using tactile sensors. In the real-world experiments, we utilize a gripper to learn screw nut twisting. Our results strike a balance between exploration and exploitation.
\end{itemize}



\section{RELATED WORK}

\subsection{Vision-based Tactile Sensor}

Tactile sensors are used to digitize contact signals in the physical world. vanilla tactile sensors, also known as haptic sensors \cite{haptic}, have been mostly based on strain principle, but their limited size makes it difficult to obtain high-resolution information. In recent years, vision-based tactile sensors such as the gelsight digit tactile sensor \cite{GelSight} have emerged. These sensors visualize the deformation of the contact surface by using a camera to convert touch into vision. Due to their high resolution and rich information, this type of sensor has unique advantages and has been successfully applied to various robot operation tasks, such as contour following \cite{contour-following}, cutting \cite{cutting}, dish loading \cite{dish-loading}, and robotic manipulation \cite{cable-manipulation}.

Based on vision-based tactile sensors, accurate depth images of the contact surface can be recovered using poisson reconstruction \cite{poisson-reconstruct}. High-resolution local contact measurements can be used to represent contact features such as the pose and force of grasping cables \cite{cable-manipulation}. Here, we use it to estimate the contact state of the manipulated object, including the position and intensity of the contact. Fig. \ref{sim_scene} shows the contact state during operation.
In addition, using neural networks to process the images from tactile sensors has become a popular approach. Feature extraction using neural networks can handle more complex contact \cite{swingbot} and perform surface reconstruction of objects \cite{poisson-reconstruct}\cite{3d-reconstruct}. However, in this work, we only consider some simple contact scenarios, so we leave it to future research to explore the use of neural networks for tactile processing.

\subsection{Robotic Reinforcement Learning and Active Inference}

Many works focus on incorporating reinforcement learning into robotics to create intelligent robots with self-iterative capabilities. Previously, the behavior of robots often required manual definition by the designer, but RL enables robots to have the ability to explore and learn on their own. However, RL is often data inefficient, so to run RL driven robots with limited real-world data, sim2real or improving the performance of RL itself can be used.
Regarding sim2real, quadruped robot could train extensively on motion behavior in simulation and then transfer to the real world \cite{ETH-quadruped}\cite{sjtu-quadruped}. Meanwhile, half sim2real reduces the amount of training required on the real world by treating simulation as a pre-training parameter process.
In terms of improving the performance of RL itself, VPG \cite{VPG} achieves unstructured object placement through the selection of several predefined skills executed by the robotic arm on image pixels, where the predefined skills reduce the algorithm's exploration space, and accelerate training speed. RHER \cite{RHER} improves data utilization through hierarchical thinking and HER methods, which can be used for sparse rewards. The advantage of using sparse rewards in robot reinforcement learning is that the reward function defined based on behavior can be complex and difficult to adjust, while the success or failure of the task is easy to evaluate.

Active inference \cite{active-inference} arises from the free energy principle\cite{friston_free_energy}. Originally it used to describe behavioral motivation in biological organisms, the free energy principle states that organisms tend to spontaneously decrease their free energy while operating, which can be decomposed into accurate understanding of the world model and action towards desired states. To simulate this process of decreasing free energy, active inference can be used for variational optimization. Active inference can be naturally extended to intelligent agents such as robots that have actuators. For example, active inference adaptive control \cite{AIC}\cite{AIC2} does not require knowledge of the robot's geometric structure and can achieve smooth multi-joint control during the dynamic reduction of free energy.  However, it have not attempted to precisely construct an external world model of the intelligent agent, but rather use variational free energy minimization to approach the desired state. 
An approach related to our work is the Active Inference Reinforcement Learning (AIRL) framework \cite{airl}, which interprets active inference in the context of RL, where free energy includes both modeling accuracy and reward.

Improving the efficiency of robot skill learning is a major challenge in the widespread use of robots. We believe that a viable solution is to use active inference, which can make full use of the data generated by the robot's interactions in the physical world and generate as much useful data as possible. In this work, we employ active inference reinforcement learning to improve the algorithm's exploratory performance and data utilization performance.

\section{METHOD}

\begin{figure}[thpb]
      \centering
      \includesvg[scale=0.4]{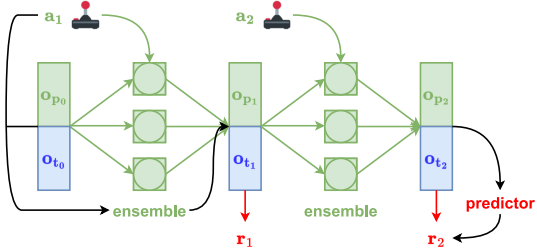}
      \caption{
      Model training loop. The green lines represent the model calculation, while the black lines depict model updates based on the loss between the data and the evaluated model. Here, $o_{p_i}$ represents the state vector of the manipulator and objects, $o_{t_i}$ represents the tactile features, $a_i$ denotes the action, $r_i$ denotes the reward, and $c_i$ represents the curiosity item at each time step. The \textbf{ensemble} and \textbf{predictor} refer to the neural network models to be learned.
      }
      \label{train}
    \end{figure}

\begin{figure}[thpb]
      \centering
      \includesvg[scale=0.4]{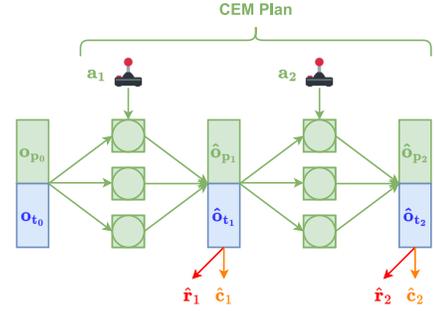}
      \caption{Planning Method. The planning method involves predicting future observations, rewards, and curiosity terms. Subsequently, the Cross-Entropy Method (CEM) is utilized to generate near-optimal actions. It is worth noting that the symbol "hat" denotes an estimate, whereas the symbol without the "hat" represents the actual data.}
      \label{run}
    \end{figure}

In this section, we introduce our Tactile-AIRL, an efficient learning algorithm that incorporates tactile information during manipulation tasks and trains agents based on the active inference principle. When it comes to robot RL, improving learning efficiency and rewards design are crucial. The major issues lie in the limited efficiency of physical sampling and the difficulties in defining rewards for complex manipulation behaviors. Therefore, the proposed pipeline aims to train manipulation skills with minimal sampling and to adapt to sparse rewards, thereby increasing the practicality of skill training and reducing the difficulty of reward function design.
The overall workflow of Tactile-AIRL is presented in Figure \ref{train} and Figure \ref{run}.

\subsection{Active Inference Reinforcement Learning}

The active inference of the agent aims to minimize its own free energy during action, leading to a more precise understanding of the agent's environment. Additionally, the agent's observations will align better with its prior knowledge. To accomplish this, we employ the decision-making scheme known as Free Energy of the Expected Future (FEEF) \cite{airl}. By considering the minimization of current and future free energy, this approach transforms the task into a planning problem, thereby accelerating learning and enhancing the stability of the current step's execution.


Let the variable $x_{t:T}$ denote a sequence of variables over time, $x_{t:T} = {x_t, \cdots ,x_T}$. Here, $\theta$ represents the parameters of the neural networks, and $\pi$ represents the policy. Next, we define the concatenation of $o_{p_i}$ and $o_{t_I}$ as $o_i$, and consider the reward as a partial observation.
Furthermore, $q(r_{t:T}, o_{t:T},\theta, \pi)$ represents a robot's beliefs regarding future variables, while $p^\Phi(r_{t:T}, o_{t:T}, \theta)$ represents a biased generative model for the robot. The FEEF (Free Energy of Expected Future) that needs to be minimized is defined as follows:

\begin{equation}
\label{FEEF}
\widetilde{\mathcal{F}}=
D_{KL}\Bigl(
q(r_{0:T},o_{0:T},\theta,\pi)||p^\Phi(r_{0:T},o_{0:T},\theta)
\Bigr)
\end{equation}

Noting that the FEEF contains the policy $\pi$, we will minimize the FEEF by adjusting the policy $q(\pi)$. After some derivation, we can get:

\begin{equation}
\widetilde{\mathcal{F}}=0 \Rightarrow D_{KL}\Bigl(q(\pi)||e^{-\widetilde{\mathcal{F}}_\pi}\Bigr)=0
\end{equation}

where
\begin{equation}
\widetilde{\mathcal{F}}_\pi=D_{KL}\Bigl(q(r_{0:T},o_{0:T},\theta|\pi)||p^\Phi(r_{0:T},o_{0:T},\theta)\Bigr)
\end{equation}

So, we can fit the distribution $q(\pi)$ to $e^{-\widetilde{\mathcal{F}}_\pi}$ to obtain a policy that minimizes free energy. Moreover, $q(\pi)$ will become more stable when $\widetilde{\mathcal{F}}_\pi$ is minimized to zero. Consequently, the problem at hand exhibits similarities with model-based reinforcement learning. By utilizing a planning algorithm to minimize the future $\widetilde{\mathcal{F}}_\pi$, we can derive a near-optimal policy.

Minimizing $\widetilde{\mathcal{F}}_\pi$ holds practical significance. To simplify the notation, let's temporarily abbreviate the sequences $r_{0:T}$ and $o_{0:T}$ as $r$ and $s$, respectively. We can decompose $\widetilde{\mathcal{F}}_\pi$ into two terms: the expected information gain term and the extrinsic term, as follows:

\begin{equation}
\label{split_free_energy}
\begin{aligned}
-\widetilde{\mathcal{F}}_\pi\approx&\mathbb{E}_{q(o,\theta|r,\pi)q(r|\pi)}\Bigl[\ln p(o,\theta|r,\pi)  - \ln q(o,\theta|\pi)  \Bigr]   \\ & -\mathbb{E}_{q(r|o,\theta,\pi)q(o,\theta|\pi)}\Bigl[ \ln q(r|o,\theta,\pi) - \ln p^\Phi(r)   \Bigr]\\
=&\underbrace{\mathbb{E}_{q(r|\pi)}\Bigl[D_{KL}\Bigl(q(o,\theta|r,\pi) || q(o,\theta|\pi)   \Bigr)\Bigr]}_{\text{expected information gain term, c}}  \\
&-\underbrace{\mathbb{E}_{q(o,\theta|\pi)}\Bigl[D_{KL}\Bigl( q(r|o,\theta,\pi)|| p^\Phi(r) \Bigr)\Bigr]}_{\text{extrinsic term, r}}
\end{aligned}
\end{equation}

Thus, the minimization of $\widetilde{\mathcal{F}}_\pi$ is equivalent to the maximization of expected information gain, indicating a preference for observations that provide new information. This intrinsic curiosity bears resemblance to mutual information \cite{aerm}. Simultaneously, minimizing $\widetilde{\mathcal{F}}_\pi$ also reduces the extrinsic term, aiming to bring future observations closer to prior observations. In the context of AIRL, this preference for priors is achieved through the expected value of the reward function. Consequently, the process of minimizing $\widetilde{\mathcal{F}}_\pi$ serves a dual role in both exploration and exploitation.

More precisely, the execution of AIRL can be divided into three steps: evaluating the future states, evaluating $\widetilde{\mathcal{F}}_\pi$, and learning the policy.

\subsubsection{Evaluating the Future States}

The reason for evaluating future states is that calculating $\widetilde{\mathcal{F}}_\pi$ requires a distribution of future states over a period of time, which can be predicted from the current time using the trained state transition ensemble model:

\begin{equation}
\begin{aligned}
\label{every_step}
q\left(o_{t: T}, r_{t: T}, \theta \mid \pi\right) & =p(\theta) 
\hookleftarrow \\ 
\hookrightarrow \prod_{\tau=t}^T &q\left(r_\tau \mid o_\tau, \theta, \pi\right) q\left(o_\tau \mid o_{\tau-1}, \theta, \pi\right) \\
q\left(r_\tau \mid o_\tau, \theta, \pi\right) & =\mathbb{E}_{q\left(o_\tau \mid \theta, \pi\right)}\left[p\left(r_\tau \mid o_\tau\right)\right] \\
q\left(o_\tau \mid o_{\tau-1}, \theta, \pi\right) & =\mathbb{E}_{q\left(o_{\tau-1} \mid \theta, \pi\right)}\left[p\left(o_\tau \mid o_{\tau-1}, \theta, \pi\right)\right]
\end{aligned}
\end{equation}

\subsubsection{Evaluating $\widetilde{\mathcal{F}}_\pi$}

Estimating $\widetilde{\mathcal{F}}_\pi$ involves evaluating the state trajectory generated by a specific action sequence during action sampling. Subsequently, the trajectory with a lower free energy will be selected. After some derivation, $\widetilde{\mathcal{F}}_\pi$ at each time step is calculated as follows. The summation of these values in future steps can be regarded as the value of future actions.

\begin{equation}
\begin{aligned}
&-\widetilde{\mathcal{F}}_{\pi_\tau}\approx -\mathbb{E}_{q(o_\tau,\theta|\pi)}\Bigl[D_{KL}\Bigl( q(r_\tau|o_\tau,\theta,\pi)||p^\Phi(r_\tau)  \Bigr)\Bigr]\\
&+\underbrace{\textbf{H}\Bigl[\mathbb{E}_{q(\theta)}[q(o_\tau|o_{\tau-1},\theta,\pi)]\Bigr] - \mathbb{E}_{q(\theta)}\Bigl[\textbf{H}[q(o_\tau|o_{\tau-1},\pi,\theta)]   \Bigr]}_{\text{state information gain, c}}
\end{aligned}
\end{equation}

\subsubsection{Learning Policy}

Because we can only obtain a score (free energy) corresponding to a certain sampled action sequence, optimization-based planning methods are not appropriate. In this work, we choose the sampling-based CEM planning algorithm. The iterative target value is $-\widetilde{\mathcal{F}}_\pi$, which is represented numerically as $e^{-\widetilde{\mathcal{F}}_\pi}$. At each action execution, a Gaussian distribution of action sequences is initialized as the initial value for planning actions. Then, the Gaussian distribution parameters are updated by sampling actions and evaluating $\widetilde{\mathcal{F}}_\pi$. Once the parameters of the Gaussian distribution stabilize, the first action is taken as the executed action.

In the above process, establishing a state transition model is similar to model-based reinforcement learning, which plays a major role in improving data efficiency. The term "information gain" in free energy, similar to intrinsic curiosity in reinforcement learning, plays a major role in improving exploratory performance.

\subsection{Tactile Information}

Vision-based tactile sensors provide a detailed depiction of contact in manipulation tasks. In this study, we utilize tactile information to complement the state space of reinforcement learning, thereby achieving superior performance in complex manipulation tasks compared to relying solely on manipulator pose or camera images. Specifically, we incorporate pose information and numerical features extracted from tactile images into the state space, with different tasks requiring distinct tactile features. This approach can also be regarded as multimodal as it combines both geometry and tactile senses as sources of information.

Commonly used tactile features encompass the centroid and summation, which are computed from depth images capturing contact information. Depth images can be obtained by applying the Poisson integral to vertical surface gradients \cite{gelsight_depth}. Additionally, the displacement field of the tactile surface is utilized to quantify object shear, which can be evaluated through optical flow analysis of RGB images. The calculation details of these features are described below.

For a simple contact, such as an object with a regular-shaped surface in contact with the tactile sensor, depth images usually consist of single connected areas. Therefore, the calculation can be performed on the complete depth images. In this study, we focus on this simple case. However, for complex contacts, it is necessary to first identify the effective depth area, which can also be considered as the region of interest. Suppose the depth image's (i,j)-th raw moments of the area to be calculated are denoted as $m_{ij}$. The centroid and summation can be obtained as follows:

\begin{align}
\mu&=(m_{10}/m_{00}, m_{01}/m_{00})\\
\Sigma&=m_{00}
\end{align}

where $\mu$ is centroid of tactile depth images, and $\Sigma$ is the sum of tactile depth images.

\begin{figure}[thpb]
      \centering
      \includegraphics[scale=0.43]{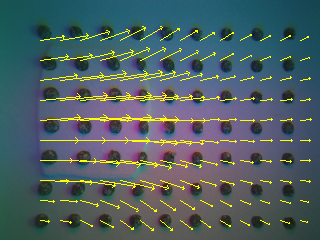}
      \caption{Optical flow when the object has a tendency to move to the right. The yellow arrow indicates the vector of optical flow for the sampled points. The entropy of optical flow distributions is considered as an estimation of shear.}
      \label{opt_flow}
    \end{figure}

The two features mentioned above describe the vertical state of contact, while the horizontal state (shear) can be characterized by gel slip. There are two approaches to estimating gel slip: gel marker point matching and optical flow method. In this study, we employ the Lucas-Kanade method to calculate the optical flow field, as shown in Fig. \ref{opt_flow}. Subsequently, we determine its distribution probability based on the calculated values and assess its distribution entropy in two directions, which indicates the magnitude of shear \cite{gelsight_opt_flow}. A higher entropy of optical flow corresponds to a greater shear amplitude caused by objects. This process can be formalized as follows:

\begin{equation}
\begin{aligned}
f&=\{(x_1,y_1), (x_2,y_2), \cdots (x_n,y_n)\}\\
H(x)&=-\int_xp(x)\log p(x)dx\\
H(y)&=-\int_yp(y)\log p(y)dy\\
\end{aligned}
\end{equation}

Thus, the tactile features $o_t$ contains: $\mu$, $\Sigma$, $H(x)$ and $H(y)$.

\section{EXPERIMENTS}

\subsection{Task}

To validate our algorithm and compare it with other methods, we focus on robot manipulation tasks in this work. Since human knowledge cannot fully encompass dynamic and unstructured environments, decision planning using rule-based control in robot manipulation tasks presents significant challenges. Therefore, robot manipulation tasks provide an ideal scenario for the application of artificial intelligence methods. Specifically, we have established robot manipulation tasks in both simulation and physical environments.
In the simulation task, we utilize tactile information to push a ball or a box on a slope. This task allows us to evaluate the learning efficiency, adaptability to different geometries, and exploration capabilities of our algorithm in scenarios with sparse rewards. Additionally, we have included a physical task involving screwing by tactile feedback, which is a common scenario in industries. Further details will be presented in subsequent sections.

\subsection{Simulator and Physical Setting}

Considering the potential risks associated with employing real robots for exploration in physical experiments, we conducted benchmarking of various approaches using robot simulation. To construct the task scenarios for robot manipulation, we utilized PyBullet and a vision-based tactile sensor plugin known as TACTO \cite{TACTO}. PyBullet serves as a Python interface for the Bullet Physics SDK, providing convenient access to Bullet functionalities for physical simulation. TACTO, which is based on the PyBullet platform, employs synchronized scene rendering of OpenGL and computes sensor outputs by parsing the contact object description files. It offers interfaces for RGB images and depth images with fast processing speed. In our simulation, we constructed a scene in PyBullet where a UR robot pushes a ball or box using the TACTO tactile sensor, as depicted in Figure \ref{sim_scene}. 

For the physical robot experiments, a robot screwing, as illustrated in Fig. \ref{real_scene}, was constructed. This scenario is commonly encountered in our daily lives. The main challenge of this task lies in the unknown pitch parameter of the screw, which the robot needs to explore using its touch sense. Another issue arises from the limited accuracy of the end effector control, where each rotation of the end effector, the height only decreases by the amount equivalent to one pitch (approximately 1.5mm~2mm). This gives rise to a position repeatability error that cannot be disregarded in the context of robotics.
Consequently, employing improper strategies may result in a mismatch between the tactile feedback and the screw nut.  This discrepancy will be reflected in the tactile feedback, thus prompting reinforcement learning for a more effective twisting strategy, even in the presence of motion deviations. Our physical robot experiment scene was implemented on the ROS2 Humble system, utilizing the KUKA iiwa7 manipulator, BackYard gripper, and GelSight mini tactile sensor.

\subsection{Environment Implementation Details}

A key element of reinforcement learning is the interaction of dataflow. In this section, we provide a detailed description of our dataflow configuration, which is formalized using the Gym wrapper.

The state vector observation, denoted as $o_p$, in the simulation consists of two parts: the position, velocity, posture, and angular velocity of the manipulator and the object to be manipulated. Moreover, since the object posture remains symmetrical during ball pushing, there is no need to consider it in this scenario. So the action $a$ involves three degrees of freedom for incremental motion: moving forward, left/right, and rotation.
And for physical robots, we have limited the movements to only two: descent and rotation around the descent direction. This reduction aims to minimize the space required for exploration and learning. Therefore, in this case, $o_p$ includes only the descent height and rotation angle. Here we enable the end effector to move downward at a fixed speed, while the action $a$ is rotation incremental motion.
The tactile features, denoted as $o_t$, in the simulation include the $\mu$ and $\Sigma$ of contact depth images, which provide a comprehensive description of the pushing object. Additionally, in the physical robot, the $H(x)$ and $H(y)$, is included in $o_t$ to capture the movement trend of the sliding nut.

Reward is usually the key determinant of the success of a reinforcement learning algorithm. In the simulation task, we have devised two reward configurations. The sparse reward configuration assigns a reward of 1.0 when the ball or box reaches the goal region, while the dense reward configuration incorporates the negative distance between the object and the center of the goal region additionally. For physical robots, our objective is to achieve an appropriate angular speed for the rotation of the screw nut and minimize the shear force exerted on the screw nut in the vertical direction. Unfortunately, obtaining the ground truth angle of the nut is challenging. Therefore, we employ the entropy of optical flow as a suitable substitute. The reward is formulated as the negative entropy in the downward direction, aiming to minimize deformation in this particular direction.



\subsection{Result and Discussion}
In the simulation, we compared our method with the vanilla RL method, Soft Actor-Critic (SAC). We conducted simulations with dense and sparse rewards, and our method outperformed SAC in terms of average reward and training episodes. However, when it came to physical robots, we only implemented our method due to limitations posed by sampling costs. To mitigate data fluctuations, we computed rewards based on a sliding window of 10 episodes for simulations and 5 episodes for experiments. Additionally, we ran our method three times in the simulation using different random seeds to plot the mean and variance.

\subsubsection{Simulation with Dense Reward}

\begin{figure}[thpb]
      \centering
      \includegraphics[scale=0.45]{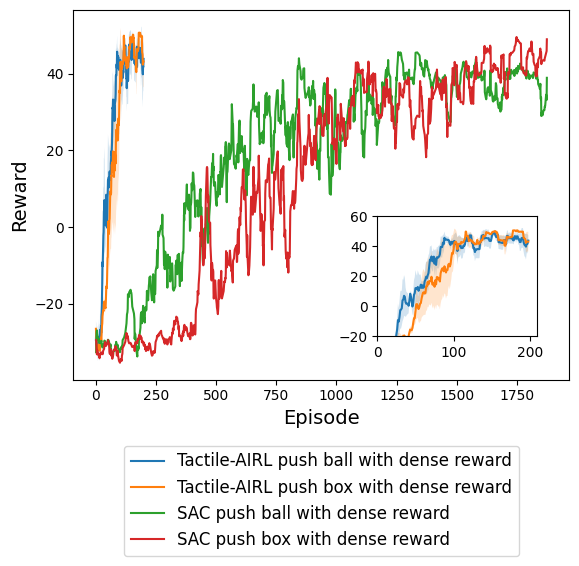}
      \caption{Experiments in dense reward configuration. Compare the use of our method and SAC for pushing ball and pushing box.}
      \label{dense_exp}
    \end{figure}

In the dense reward configuration, our method and most RL methods can learn the skill of pushing the ball. As shown in the comparison results in Figure \ref{dense_exp}, we found that Tactile-AIRL was an order of magnitude more effective than the RL baseline. Tactile-AIRL achieved almost maximum episode rewards by interacting with the environment for only 100 episodes, while SAC took about 1000 episodes. The reason behind this improvement is that we introduce tactile information and employ a modeling approach similar to model-based reinforcement learning to extract information from existing data, thereby significantly enhancing data efficiency.


\subsubsection{Simulation with Sparse Reward}

\begin{figure}[thpb]
\vspace{0.1cm}
      \centering
      \includegraphics[scale=0.45]{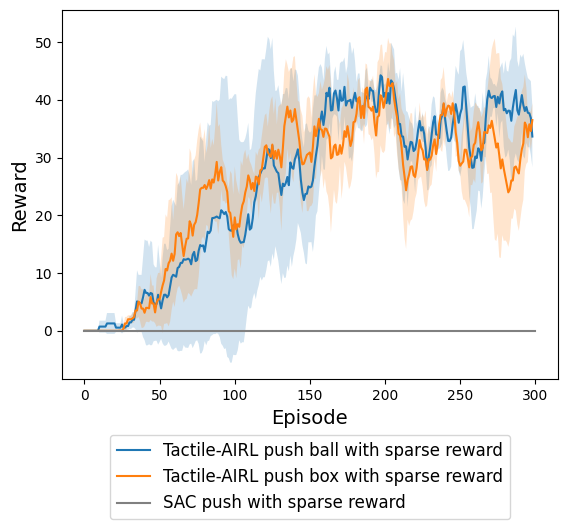}
      \caption{Experiments in Sparse Reward Configuration. In this case, it has been observed that SAC fails to learn skills. }
      \label{sparse_exp}
    \end{figure}

In the sparse reward configuration, most RL methods cannot learn the skill of pushing. As depicted in Figure \ref{sparse_exp}, we observed that the RL baseline failed to learn even after exploring thousands of episodes, whereas our method consistently achieved high performance.  This capability arises from the algorithm's objective of minimizing free energy, which drives it to acquire new information during exploration until it discovers states with effective rewards. Thus, our method demonstrates advantages in learning robotic manipulation skills by alleviating the burden of reward design.

\subsubsection{Physical Experiment}

\begin{figure}[thpb]
      \centering
      \includegraphics[scale=0.4]{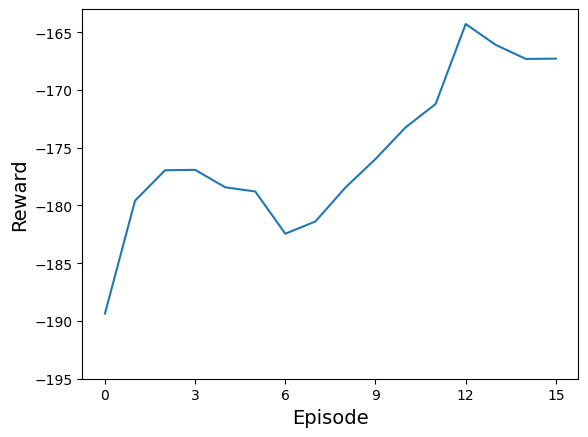}
      \caption{Experiments in physical robot. Achieve the skill of minimizing downward shear force in relatively few episodes. }
      \label{real_exp}
    \end{figure}

In order to accelerate learning in the real world and avoid the need for precise measurement of screw ground truth, we currently refrain from using sparse rewards. Instead, we employ dense rewards based on shear estimation. As depicted in Figure \ref{real_exp}, our approach demonstrates convergence within acceptable training periods, providing strong evidence for the data efficiency of our algorithm in real-world applications. The incorporation of tactile sensation enables us to estimate the effectiveness of task execution and enhances the perception space. Moreover, the utilization of active inference reinforcement learning expedites skill acquisition, highlighting the potential of our algorithm for broader application scenarios.

\section{CONCLUSIONS}

We propose an efficient learning method for robot manipulation skills, called Tactile-AIRL, which uses active inference RL to train robot agents in environments equipped with tactile sensing. The method encourages agents to explore effectively and utilizes the tactile data from exploration interactions for model learning, synchronously improving both exploration and exploitation. Experiments shows that Tactile-AIRL has unique advantages in the field of robotics.

One limitation of this work is that the hand-crafted features of the tactile sensor prevent our method from achieving higher generalization. Therefore, we also plan to incorporate multimodal reinforcement learning in future work to improve the generalizability of our method to manipulation tasks. This can be achieved by directly taking tactile images or scene images as input and fusing them with a state vector. We believe that this approach will further advance the use of vision-based tactile sensors in manipulation learning, which will be explored in future work.








\bibliographystyle{IEEEtranBST/IEEEtran}
\bibliography{IEEEtranBST/IEEEabrv,ref}

\begin{thebibliography}{10}
\providecommand{\url}[1]{#1}
\csname url@rmstyle\endcsname
\providecommand{\newblock}{\relax}
\providecommand{\bibinfo}[2]{#2}
\providecommand\BIBentrySTDinterwordspacing{\spaceskip=0pt\relax}
\providecommand\BIBentryALTinterwordstretchfactor{4}
\providecommand\BIBentryALTinterwordspacing{\spaceskip=\fontdimen2\font plus
\BIBentryALTinterwordstretchfactor\fontdimen3\font minus
  \fontdimen4\font\relax}
\providecommand\BIBforeignlanguage[2]{{%
\expandafter\ifx\csname l@#1\endcsname\relax
\typeout{** WARNING: IEEEtran.bst: No hyphenation pattern has been}%
\typeout{** loaded for the language `#1'. Using the pattern for}%
\typeout{** the default language instead.}%
\else
\language=\csname l@#1\endcsname
\fi
#2}}

\bibitem{tactile-servo}
A.~Delgado, C.~Jara, and F.~Torres, ``In-hand recognition and manipulation of
  elastic objects using a servo-tactile control strategy,'' \emph{Robotics and
  Computer-Integrated Manufacturing}, vol.~48, pp. 102--112, 2017.

\bibitem{haptic-touch}
N.~Jamali, C.~Ciliberto, L.~Rosasco, and L.~Natale, ``Active perception:
  Building objects' models using tactile exploration,'' in \emph{2016 IEEE-RAS
  16th International Conference on Humanoid Robots (Humanoids)}, 2016, pp.
  179--185.

\bibitem{GelSight}
R.~Li, R.~Platt, W.~Yuan, A.~ten Pas, N.~Roscup, M.~A. Srinivasan, and
  E.~Adelson, ``Localization and manipulation of small parts using gelsight
  tactile sensing,'' in \emph{2014 IEEE/RSJ International Conference on
  Intelligent Robots and Systems}, 2014, pp. 3988--3993.

\bibitem{muzero}
J.~Schrittwieser, I.~Antonoglou, T.~Hubert, K.~Simonyan, and L.~Sifre,
  ``Mastering atari, go, chess and shogi by planning with a learned model,''
  \emph{Nature}, vol. 588, no. 7839, pp. 604--609, Dec 2020.

\bibitem{dreamer}
D.~Hafner, T.~Lillicrap, J.~Ba, and M.~Norouzi, ``Dream to control: Learning
  behaviors by latent imagination,'' \emph{arXiv preprint arXiv:1912.01603},
  2019.

\bibitem{haptic}
P.~Payeur, C.~Pasca, A.-M. Cretu, and E.~Petriu, ``Intelligent haptic sensor
  system for robotic manipulation,'' \emph{IEEE Transactions on Instrumentation
  and Measurement}, vol.~54, no.~4, pp. 1583--1592, 2005.

\bibitem{contour-following}
N.~F. Lepora, A.~Church, C.~de~Kerckhove, R.~Hadsell, and J.~Lloyd, ``From
  pixels to percepts: Highly robust edge perception and contour following using
  deep learning and an optical biomimetic tactile sensor,'' \emph{IEEE Robotics
  and Automation Letters}, vol.~4, no.~2, pp. 2101--2107, 2019.

\bibitem{cutting}
A.~Yamaguchi and C.~G. Atkeson, ``Combining finger vision and optical tactile
  sensing: Reducing and handling errors while cutting vegetables,'' in
  \emph{2016 IEEE-RAS 16th International Conference on Humanoid Robots
  (Humanoids)}, 2016, pp. 1045--1051.

\bibitem{dish-loading}
N.~Kuppuswamy, A.~Alspach, A.~Uttamchandani, S.~Creasey, T.~Ikeda, and
  R.~Tedrake, ``Soft-bubble grippers for robust and perceptive manipulation,''
  in \emph{2020 IEEE/RSJ International Conference on Intelligent Robots and
  Systems (IROS)}, 2020, pp. 9917--9924.

\bibitem{cable-manipulation}
Y.~She, S.~Wang, S.~Dong, N.~Sunil, A.~Rodriguez, and E.~Adelson, ``Cable
  manipulation with a tactile-reactive gripper,'' \emph{The International
  Journal of Robotics Research}, vol.~40, no. 12-14, pp. 1385--1401, 2021.

\bibitem{poisson-reconstruct}
S.~Wang, J.~Wu, X.~Sun, W.~Yuan, W.~T. Freeman, J.~B. Tenenbaum, and E.~H.
  Adelson, ``3d shape perception from monocular vision, touch, and shape
  priors,'' in \emph{2018 IEEE/RSJ International Conference on Intelligent
  Robots and Systems (IROS)}, 2018, pp. 1606--1613.

\bibitem{swingbot}
C.~Wang, S.~Wang, B.~Romero, F.~Veiga, and E.~Adelson, ``Swingbot: Learning
  physical features from in-hand tactile exploration for dynamic swing-up
  manipulation,'' in \emph{2020 IEEE/RSJ International Conference on
  Intelligent Robots and Systems (IROS)}, 2020, pp. 5633--5640.

\bibitem{3d-reconstruct}
E.~Smith, R.~Calandra, A.~Romero, G.~Gkioxari, D.~Meger, J.~Malik, and
  M.~Drozdzal, ``3d shape reconstruction from vision and touch,'' in
  \emph{Advances in Neural Information Processing Systems}, H.~Larochelle,
  M.~Ranzato, R.~Hadsell, M.~Balcan, and H.~Lin, Eds., vol.~33.\hskip 1em plus
  0.5em minus 0.4em\relax Curran Associates, Inc., 2020, pp. 14\,193--14\,206.

\bibitem{ETH-quadruped}
T.~Miki, J.~Lee, J.~Hwangbo, L.~Wellhausen, V.~Koltun, and M.~Hutter,
  ``Learning robust perceptive locomotion for quadrupedal robots in the wild,''
  \emph{Science Robotics}, vol.~7, no.~62, p. eabk2822, 2022.

\bibitem{sjtu-quadruped}
J.~Wu, G.~Xin, C.~Qi, and Y.~Xue, ``Learning robust and agile legged locomotion
  using adversarial motion priors,'' \emph{IEEE Robotics and Automation
  Letters}, vol.~8, no.~8, pp. 4975--4982, 2023.

\bibitem{VPG}
A.~Zeng, S.~Song, S.~Welker, J.~Lee, A.~Rodriguez, and T.~Funkhouser,
  ``Learning synergies between pushing and grasping with self-supervised deep
  reinforcement learning,'' in \emph{2018 IEEE/RSJ International Conference on
  Intelligent Robots and Systems (IROS)}, 2018, pp. 4238--4245.

\bibitem{RHER}
Y.~Luo, Y.~Wang, K.~Dong, Q.~Zhang, E.~Cheng, Z.~Sun, and B.~Song, ``Relay
  hindsight experience replay: Self-guided continual reinforcement learning for
  sequential object manipulation tasks with sparse rewards,''
  \emph{Neurocomputing}, vol. 557, p. 126620, 2023.

\bibitem{active-inference}
F.~R. Karl~Friston, Thomas~FitzGerald, P.~Schwartenbeck, J.~O'Doherty, and
  G.~Pezzulo, ``Active inference and learning,'' \emph{Neuroscience \&
  Biobehavioral Reviews}, vol.~68, pp. 862--879, 2016.

\bibitem{friston_free_energy}
K.~Friston, ``The free-energy principle: a unified brain theory?'' \emph{Nature
  Reviews Neuroscience}, vol.~11, no.~2, pp. 127--138, Feb 2010.

\bibitem{AIC}
C.~Pezzato, R.~Ferrari, and C.~H. Corbato, ``A novel adaptive controller for
  robot manipulators based on active inference,'' \emph{IEEE Robotics and
  Automation Letters}, vol.~5, no.~2, pp. 2973--2980, 2020.

\bibitem{AIC2}
G.~Oliver, P.~Lanillos, and G.~Cheng, ``An empirical study of active inference
  on a humanoid robot,'' \emph{IEEE Transactions on Cognitive and Developmental
  Systems}, vol.~14, no.~2, pp. 462--471, 2022.

\bibitem{airl}
A.~Tschantz, B.~Millidge, A.~K. Seth, and C.~L. Buckley, ``Reinforcement
  learning through active inference,'' \emph{arXiv preprint arXiv:2002.12636},
  2020.

\bibitem{aerm}
T.~Schneider, B.~Belousov, G.~Chalvatzaki, D.~Romeres, D.~K. Jha, and
  J.~Peters, ``Active exploration for robotic manipulation,'' in \emph{2022
  IEEE/RSJ International Conference on Intelligent Robots and Systems (IROS)},
  2022, pp. 9355--9362.

\bibitem{gelsight_depth}
S.~Wang, Y.~She, B.~Romero, and E.~Adelson, ``Gelsight wedge: Measuring
  high-resolution 3d contact geometry with a compact robot finger,'' in
  \emph{2021 IEEE International Conference on Robotics and Automation (ICRA)},
  2021, pp. 6468--6475.

\bibitem{gelsight_opt_flow}
W.~Yuan, R.~Li, M.~A. Srinivasan, and E.~H. Adelson, ``Measurement of shear and
  slip with a gelsight tactile sensor,'' in \emph{2015 IEEE International
  Conference on Robotics and Automation (ICRA)}, 2015, pp. 304--311.

\bibitem{TACTO}
S.~Wang, M.~Lambeta, P.-W. Chou, and R.~Calandra, ``{TACTO}: A fast, flexible,
  and open-source simulator for high-resolution vision-based tactile sensors,''
  \emph{IEEE Robotics and Automation Letters (RA-L)}, vol.~7, no.~2, pp.
  3930--3937, 2022.

\end{thebibliography}
\end{document}